\pdfoutput=1

\documentclass[11pt]{article}

\usepackage[]{acl}

\usepackage{times}
\usepackage{latexsym}
\usepackage{CJKutf8}
\usepackage[T1]{fontenc}
\usepackage{amsmath, bm}
\usepackage{amsfonts,amssymb}
\usepackage{multirow}
\usepackage{graphicx}
\usepackage{microtype}
\usepackage{amsfonts}
\usepackage{mathrsfs}
\usepackage{multirow}
\usepackage{amssymb}
\usepackage{subfigure} 
\usepackage{CJKutf8}
\usepackage{colortbl}
\usepackage{makecell}

\usepackage[utf8]{inputenc}

\usepackage{microtype}

\title{Chinese Word Segmentation with Heterogeneous Graph Neural Network}

\author{Xuemei Tang$^1$, {\bf Jun Wang$^1$}, {\bf Qi Su$^{2*}$} \\
        $^{1}$ Department of Information Management, Peking University \\
        $^{2}$ School of Foreign Language, Peking University\\
        \texttt{tangxuemei@stu.pku.edu.cn,\{junwang,sukia\}@pku.edu.cn}}

\begin{document}
\maketitle
\begin{CJK*}{UTF8}{gbsn}

\begin{abstract}
In recent years, deep learning has achieved significant success in the Chinese word segmentation (CWS) task. Most of these methods improve the performance of CWS by leveraging external information, e.g., words, sub-words, syntax. However, existing approaches fail to effectively integrate the multi-level linguistic information and also ignore the structural feature of these external information. Therefore, in this paper, we proposed a framework to improve CWS, named HGNSeg. It exploits multi-level external information sufficiently with the pre-trained language model and heterogeneous graph neural network. The experimental results on six benchmark datasets (e.g., Bakeoff 2005, Bakeoff 2008) validate that our approach can effectively improve the performance of Chinese word segmentation. Importantly, in cross-domain scenarios, our method also shows a strong ability to alleviate the OOV problem.
\end{abstract}

\section{Introduction}

Unlike English and France with space between words, many Asian languages don't contain delimiters such as spaces to mark the boundaries of words, like Chinese. Therefore, Chinese word segmentation is generally considered as an important step in the Chinese natural language processing (NLP) pipeline, providing information to downstream tasks, such as information extraction and question answering.

\begin{figure*}[t]
\centering
\includegraphics[height=5cm,width=14cm]{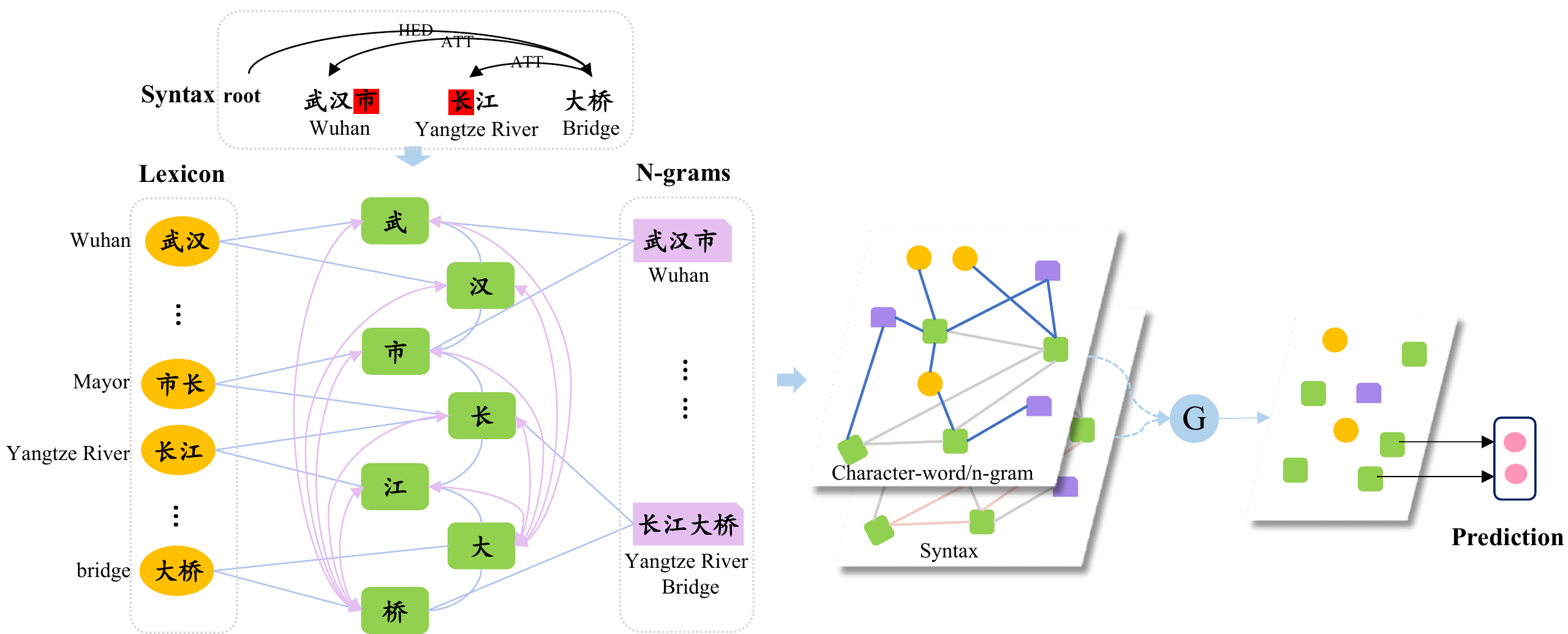}

\caption{The architecture of HGNSeg. Blue lines represent the  character-word/n-gram edges, pink curves represent the dependency syntactic edges.}
\label{p1}
\end{figure*}

Recently, many neural network approaches are proved to be successful in CWS \citep{em:7,em:8,em:9,em:20,em:17,em:36,em:37,em:34,em:35,em:27,em:29,em:42,em:30,em:31,em:26,em:24}, where various methods were proposed to incorporate the context information and the external resource to improve the performance of CWS \citep{em:37,em:36,em:35,em:34,em:30,em:31,em:26,em:1,em:51}. However, those methods lack explicit model to make full use of the syntax and multi-level linguistic information. As we know, the dependency syntactic structure is also able to provide the information of the relationship between words, also can help relieve ambiguity segmentation. For example, the sentence given in  Figure~\ref{p1} can be segmented as ``武汉市(Wuhan)/长江(Yangtze River)/大桥(Bridge)'', or ``武汉(Wuhan)/市长(Mayor)/江大桥(Jiang Daqiao)'', the dependency syntactic structure could capture the correct syntactic relationship in this sentence, ``武汉(Wuhan)'' depends on ``大桥(Bridge)''. Although some works tried to use the dependency syntactic tree information \citep{em:31}, they neglected the information of tree structure. One potential reason is the lack of simple and effective methods for incorporating structural information into neural encoders. 

Recently, a new neural network called graph neural network (GNN) has attracted wide attention \citep{em:79,em:55}. Graph neural networks have been effective at tasks thought to have rich relational structure and can preserve global structure information of a graph in graph embeddings \citep{em:57}. It has been applied in several NLP tasks to encode external knowledge with complex structure successfully \citep{em:53,em:54,em:60}.

In this work, we propose a multi-level features learning framework for CWS with a heterogeneous graph neural network (HGNN), named HGNSeg. Specifically, we incorporate the multi-level linguistic features (e.g., words, n-grams, dependency syntactic tree) and their structural information via HGNN, where a node takes the information from neighboring nodes to update its representation. We represent all characters in the input sentence, the possible segmented words from lexicon, high frequency n-grams as nodes, then add different types of edge into the graph. The Characters connect each other with the dependency syntactic relationship, and the character connect the word and n-gram according to its relative position in them.

To summarize, our contributions are as follows:

\begin{itemize}

\item We propose a novel graph-based framework to easily integrate heterogeneous information for CWS. To the best of our knowledge, this is the first work using the heterogeneous graph neural networks for CWS.

\item We encode multi-level linguistic features (e.g., character, word, n-gram and dependency syntactic) via combining the pre-trained language model and the heterogeneous graph neural network, in which we design multiple types of nodes and edges to integrate the structural information of these linguistic features.

\item Results on six benchmark datasets demonstrate that our model outperforms state-of-the-art CWS methods. Our method also relieve OOV issue effectively by incorporating multi-level linguistic information in the cross-domain scenario. 

\end{itemize}

\section{Related Work}

\citep{em:6} first models Chinese word segmentation as a sequence labeling task. Recently, many neural network approaches are proved to be effective in this task \cite{em:37,em:36,em:35,em:34,em:30,em:31,em:26}. Among these works, an essential direction is to explore effective methods for integrating external knowledge (e.g., dictionary, syntax) and learning contextual infromation (e.g., n-grams). The study from \citet{em:34} leveraged dictionary based on pseudo labeled data and multi-task learning to improve CWS. \citet{em:35} integrated dictionaries by constructing feature vectors for each character with pre-defined several templates. N-grams are also an rich context resource for CWS \cite{em:37,em:30,em:17}, \citet{em:26} incorporated wordwood information for neural segmenter and achieved state-of-the-art performance at that time. \citet{em:30} utilized syntax knowledge generated by existing NLP toolkits to improve the CWS and part-of-speech (POS) joint task. Those methods achieved satisfactory performance, but they ignored the structure of external knowledge.



Graph convolutional networks (GCNs) is a recent type of multi-layer neural networks operating on graphs. For each node in the graph, GCNs encode relevant information about its neighborhood to update its representation. It has been applied to many NLP tasks \citep{em:51,em:52,em:53,em:57,em:56,em:1}, such as semantic role labeling, machine translation, and Chinese word segmentation. \citeauthor{em:1} leveraged GCNs encoding multi-granularity structure information to improve the joint CWS and POS  task. \citeauthor{em:51} exploited graph neural network to learn the local composition features based domain lexicon for electronic medical record text word segmentation. However, these efforts didn't make full use of the fact that GCNs can encode the structural information.

Motivated by above studies, we use the pre-trained language model and the heterogeneous graph neural network to encode multi-level linguistic features, (e.g., character, local information), syntactic knowledge.

\section{Method}

In this section, we will introduce the HGNSeg framework, including how to construct the heterogeneous graph and to implement word segmentation using HGNN. In general, we first build a heterogeneous graph based on the corpus and selected features, and next we encode all the nodes in the graph, then we pass the encoded nodes to the  HGNN, each layer of which performs graph convolution operation on sub-graphs separated by different edge types, and then aggregates the information together. Finally, the graph neural network outputs concatenate with the outputs of the pre-trained model go through the decoder to get the final label sequence. Figue~\ref{p1} shows the overall structure of our heterogeneous graph network.

Chinese word segmentation is generally viewed as a character-based sequence labeling task. Specifically, given a sentence $ X\ = \{x_1,\ x_2,... x_T\} $, each character in the sequence is labeled as one of $\ \mathcal{L}\ =\{B, M, E, S\} $, indicating the character is in the beginning, middle, end of a word, or the character is a single-character word. CWS aims to figure out the ground truth of labels $Y^* = \{y_1^*, y_2^*, … y_{T}^*\} $:

\begin{equation}\label{eq1}
Y^\ast={arg\ max\ P\left(Y\middle| X\right)}_{Y\in \mathcal{L}^T} 
\end{equation}

\subsection{Encoding Layer}
 According to \citet{em:5}, the real factor results in the gap of CWS task is un-sufficient training instead of bad-training. So, we utilize the pre-trained language model as the shared encoder, which is pre-trained with a large number of unlabeled Chinese data.

\begin{equation}\label{eq2}
[\textbf{e}_1,...,\textbf{e}_i,...{,\textbf{e}}_T]=Encoder([x_1,...,x_i,...x_{T}])
\end{equation}
where $\textbf{e}_i$ is the representation for $x_i$ from the encoder.

\subsection{Heterogeneous Graph Neural Network}
\subsubsection{Graph construction}
Encouraged by some previous works on GCN-based CWS \citep{em:1,em:51}, we construct the graph by representing characters from corpus, words from lexicon $\mathcal{D}$, and n-grams from n-grams vocabulary $\mathcal{N}$ as graph nodes. Then three types of edges are defined in the graph according to different relationships.

\textbf{Syntax Edges.} \citet{em:52} proposed GCN to operate on directed and labeled graphs. This approach make it possible to use linguistic structures such as dependency trees. They also used edges-wise gates which let the model regulate weights of individual dependency edges.

Follow their work, in our heterogeneous graph, Character-Character edges represent syntactic relationship between two characters. We follow the convention that in dependency syntactic trees heads point to their dependents, and thus outgoing edges are used for head-to-dependent connections, and incoming edges are used for dependent-to-head connections. 

We use an example to illustrate our idea. For the input sentence show in Figure~\ref{p1}. ``大桥(big bridge)'' as a head points to the dependent ``长江(Yangtze River)'', therefore ``大(big)'' and ``桥(bridge)'' have two outgoing edges to ``长(long)'' and ``江(river)'' respectively, ``大(big)'' and ``桥(bridge)'' also have two incoming edges from ``长(long)'' and ``江(river)'' respectively.

\textbf{Character-Word/N-gram Edges.} As we know, CWS aims to find the best segment position in the input sentence. However, each character $x_i$ may have different position in each word. For example, $x_i$ may be the beginning, middle, ending of a word, or $x_i$ may be a single-character word. Different positions transfer different information. Therefore, in order to learning the local contextual information and candidate words, the edge connecting $x_i$ and word/n-gram in the sub-graph represent the $x_i$ is the beginning of the words/n-gram or the ending, where the word/n-gram is a span from sentence and can be matched in the lexicon $\mathcal{D}$ or n-grams vocabulary $\mathcal{N}$. In this type of sub-graph, we also connect each character with the next character to maintain the order of the sequence.

As show in Figure~\ref{p1}, ``长'' is connects with ``市长'' and ``长江'', two words are split from the the input sentence and matched in the lexicon $\mathcal{D}$. ``长'' is also connects with ``长江大桥'', which is contained in n-grams vocabulary $\mathcal{N}$. These two types of edge are unidirectional.

\subsubsection{Graph convolutional network}
After building the heterogeneous graph, we converted the character nodes, the word nodes, and the n-gram nodes into embeddings, which are trainable during training process. Next, we apply graph convolutional operation \citep{em:59} on the two sub-graph to calculate higher-order representations of each node with aggregated information.

Given a graph $\mathcal{G} = (\mathcal{V},\bm{\varepsilon})$, where $\mathcal{V}$ and $\bm{\varepsilon}$ represent the set of nodes and edges respectively. Let $X \in \mathbb{R}^{\mid{\mathcal{V}}|\times d}$ be a matrix containing the nodes with their features $x_v \in \mathbb{R}^d$ (each row $x_v$ is a feature vector for a node $v$). For the graph $\mathcal{G}$ ,we introduce its adjacency matrix $A^{'} = A + I$ with added self-connections and degree matrix $D$, where $D_{ii} = \sum_{j}A_{ij}^{'} $. Then the layer-wise propagation rule is as follow.

\begin{equation}
    H^{(l+1)} = \sigma(\tilde{A} \cdot H^{(l)} \cdot W^{(l)})
\end{equation}
where $\tilde{A} = D^{1/2} A^{'} D^{1/2}$ represents the symmetric normalized adjacency matrix and $W^{(l)}$ is a layer-specific trainable transformation matrix. $\sigma(\cdot)$ denotes an activation function such as RELU. $H^{(l)} \in \mathbb{R}^{\mid{\mathcal{V}}|\times d}$ denotes the hidden representations of nodes in the $l^{th}$ layer.

In our heterogeneous graph, there are different types of edges $\mathcal{T} = \{\tau_{in},\tau_{out}, \tau_{cwn}\}$, where $\tau_{in}$ represents the syntax incoming edges, $\tau_{out}$ represents the syntax outgoing edges, $\tau_{cwn}$ represents the character-word/n-gram edges. We update the $(l+1)_{th}$ layer representation of the node $H^{(l+1)}$ by aggregating the features of their neighboring nodes $H_{\tau}^{(l)}$ with different types of $\tau$.
\begin{equation}
    H^{(l+1)} = \sigma(\sum_{\tau \in \mathcal{T}}(\tilde{A}_{\tau} \cdot H_{\tau}^{(l)} \cdot W_{\tau}^{(l)}))
\end{equation}
where $\tilde{A}_{\tau}$ is a submatrix of the symmetric normalized adjacency matrix that only contains edges with $\tau$, $H_{\tau}^{(l)}$ is the feature matrix of the neighboring nodes with type $\tau$ of each node, and $W_{\tau}^{(l)}$ is a trainable parameter. $\sigma(\cdot)$ denotes an non-linear activation function. Initially, $H_{\tau}^{(0)}$ is the node feature from the pre-trained model or random initialization.

\begin{table*}[t]
\centering
\setlength{\tabcolsep}{1mm}
\begin{tabular}{|c|c|c|c|c|c|c|c|c|c|c|c|} 
\hline
\multicolumn{2}{|c|}{}                          & \multicolumn{1}{c|}{}                    & \multicolumn{1}{c|}{}                     & \multicolumn{1}{c|}{}                     & \multicolumn{1}{c|}{}                       & \multicolumn{1}{c|}{}                      & \multicolumn{1}{c|}{}        & 

\multicolumn{4}{c|}{Cross-domain}                       \\                         
\cline{9-12}

\multicolumn{2}{|c|}{Datasets}                          & \multicolumn{1}{c|}{AS}                    & \multicolumn{1}{c|}{PKU}                     & \multicolumn{1}{c|}{MSR}                     & \multicolumn{1}{c|}{CITYU}                       & \multicolumn{1}{c|}{CTB}                      & \multicolumn{1}{c|}{SXU}                     &  \multicolumn{1}{c|}{ZX}                    & 
\multicolumn{1}{c|}{PT}                    & 
\multicolumn{1}{c|}{DL}                    & 
\multicolumn{1}{c|}{DM} \\
\hline
Words & train                  & 5.4M                                     & 1.1M                                      & 2.4M                                      & 1.1M                                        & 0.6M                                       & 0.5M                                                                           &\multicolumn{4}{c|}{PKU training set}                                                                                                          \\ 
\cline{2-12}
                       & test                   & 0.1M                                     & 0.1M                                      & 0.1M                                      & 0.2M                                        &   0.1M                                      & 0.1M                                                                           & 34.0K                                                                             & 34.0K                     & 32.0K                      & 17.0K                          \\ 
\hline
Chars & train                  & 8.3M                                     & 1.8M                                      & 4.0M                                      & 1.8M                                        & 1.0M                                       & 0.8M                                                                                                                                                      &\multicolumn{4}{c|}{PKU training set}                          \\ 
\cline{2-12}
                       & test                   & 0.2M                                     & 0.2M                                      & 0.2M                                      & 0.4M                                        & 0.1M                                        & 0.2M                                      & 48.0K                                                                                                                    & 57.0K                      & 47.0K                     & 30.0K                                         \\ 
\hline
\multicolumn{2}{|c|}{OOV rate(\%)}              & \multicolumn{1}{c|}{2.2}                 & \multicolumn{1}{c|}{2.1}                  & \multicolumn{1}{c|}{1.3}                  & \multicolumn{1}{c|}{3.7}                    & \multicolumn{1}{c|}{3.8}                     & \multicolumn{1}{c|}{2.6}                  & \multicolumn{1}{c|}{2.9}                &  \multicolumn{1}{c|}{4.9} & \multicolumn{1}{c|}{2.3} & \multicolumn{1}{c|}{5.3}\\
\hline
\end{tabular}
\caption{Statistics of 10 datasets.}
\label{t1}
\end{table*}

\subsubsection{Edge-wise gating}

Uniformly absorbing information from all neighboring nodes may not be suitable for the CWS. Due to our method rely on automatically toolkits to parse dependency syntactic structure, its performance is not perfect. It is risky for downstream application to rely on a potentially wrong syntactic edges. On the other hand, each character possible accepts information from multiple candidate words, not each candidate word is a correct segmentation in the input sentence. So the corresponding nodes message in the neural network need to be down-weighted.

To solve the aforementioned problems, inspired by recent efforts \citep{em:61,em:52,em:62}. We calculate each edge node pair with a scalar gate as follow.
\begin{equation}
    g_{\tau}^{(l)} = \theta(H_{\tau}^{(l)}\cdot W_{\tau}^{(l),g} + b_{\tau}^{(l),g})
\end{equation}
where $\theta$ is the logistic sigmoid function, $W_{\tau}^{(l),g} \in \mathbb{R}^{d} $ and $ b_{\tau}^{(l),g} \in \mathbb{R}$ are a weight and a bias for the gate. Therefore, the final heterogeneous graph convolutional neural network calculation is formulated as follow.
\begin{equation}
      H^{(l+1)} = \sigma(\sum_{\tau \in \mathcal{T}}(g_{\tau}^{(l)}(\tilde{A}_{\tau} \cdot H_{\tau}^{(l)} \cdot W_{\tau}^{(l)}))) 
\end{equation}

\subsection{Decoding Layer}
There are different algorithms can be implemented as decoders, such as random conditional fields (CRF) \cite{em:10} and softmax. In our framework, we use CRF as the decoder.

In CRF layer, $P\left(Y\middle| X\right)$ in Eq.\ref{eq1} could be represented as:

\begin{equation}\label{eq3}
P(Y|X)=\frac{\emptyset(Y|X)}{\sum_{Y^\prime\in \mathcal{L}^T}{\emptyset(Y^\prime|X)}}
\end{equation}
where, $\ \ \emptyset(Y|X)$ is the potential function, and we only consider interactions between two successive labels.

\begin{equation}\label{eq4}
\emptyset(Y|X)=\ \prod_{i=2}^{T}{\sigma(X,i,y_{i-1},y_i)}
\end{equation}
\begin{equation}\label{eq5}
\sigma(\mathbf{x},i,y^\prime,y)\ =\ exp(s{(X,i)}_y+b_{y^\prime y})
\end{equation}
where $b_{y^\prime y} \in\mathbb{\rm \textbf{R}}$ is trainable parameters respective to label pair $(y^\prime,\ y)$. The score function $s(X\ ,\ i)\ \in\mathbb{R}^{\left|\mathcal{L}\right|}$ calculate the score of each lable for $i_{th}$ character:

\begin{equation}\label{eq6}
s(X\,\ i)\ ={\ \textbf{W}}_s^\top \textbf{a}_i+b_s
\end{equation}
where $\textbf{a}_i = h_i \bigoplus e_i$, which generated by concatenate $H^{(l+1)}$ and the output of the encoder $\textbf{\emph{E}}$ . $\textbf{W}_s\in\mathbb{R}^{d_a\times L}$ and $b_s\in\mathbb{R}^{\left|\mathcal{L}\right|}$ are trainable parameters.

\subsection{Lexicon and N-gram Vocabulary Construction}

To build the character-word/n-gram sub-graph, the first step is to construct the lexicon $\mathcal{D}$ and n-gram vocabulary $\mathcal{N}$, because the edges between characters and words/n-grams are built on them. In our work, $\mathcal{D}$ consists of words from training set. For $\mathcal{N}$, we use Accessor variety (AV) \citep{em:65} to extract high frequency n-grams from each corpus. AV is un-supervised method for word extraction from Chinese text collections. The accessor variety of a string $s$ of more than one character is defined as follow.

\begin{equation}
    \emph{AV(s)} = min\{L_{av}(s),R_{av}(s)\}
\end{equation}
here $L_{av}(s)$ is called the left accessor variety and is defined as the number of distinct characters (predecessors) except the start of a sentence that precede $s$ plus the number of distinct sentences of which $s$ appears at the beginning. The right accessor variety $R_{av}(s)$ is similar to the $L_{av}(s)$. 


\section{Experiment}

\subsection{Datasets and Experimental Setup}

We experiment on six benchmark CWS datasets, including PKU, MSR, CITYU, and AS from SIGHAN 2005 \citep{em:22}, CTB and SXU from SIGHAN 2008 bake-off task \citep{em:66}. And we also run our model on four cross-domain datasets, including two Chinese fantasy novel dataset: DL (\emph{DoLuoDaLu}), ZX (\emph{ZhuXian}) \cite{em:63}, one medicine dataset: DM (dermatology), and a patent dataset: PT \cite{em:76}. The sizes of 10 corpora are shown in Table ~\ref{t1}. We randomly pick 10\% sentences from the training data as the development data for model tuning for SIGHAN 2005. Follow to a previous paper \citep{em:2}, we replace all digits, punctuation, and Latin letters to 
a unique token, dealing with the full/half-width mismatch between training and test data. AS and CITYU are mapped from traditional Chinese to simplified Chinese before segmentation. Note that we use the newswire dataset, PKU, as the source domain data for the cross-domain experiment, which is different from four target datasets.


For the encoders, we follow the the default setting of the BERT \citep{em:47} and the RoBERTa \cite{em:58}. We represent all the character nodes using the pre-trained language model. And we initialize word nodes and n-gram nodes by random initialization. To obtain the aforementioned dependency structure, we use two automatic toolkits, Stanford CoreNLP Toolkit (SCT)\footnote{\url{https:// stanfordnlp.github.io/CoreNLP/}}\citep{em:73} and LTP 4.0 \footnote{\url{https://github.com/HIT-SCIR/ltp}} \citep{em:74}. For convolutional operation, we use two-layer graph convolutional network. We set the maximum length and minimum frequency of n-gram to 5 when use AV extract n-grams from each corpus.

\subsection{Results on Benchmark Datasets}

The experimental results on the six benchmark datasets are shown in Table \ref{t3}, where the reported F1 scores and OOV recall are the average of the three-times experimental results. We list experimental results using different encoders, and to further illustrate the validity of our model, we also compare our best results on same datasets with some previous state-of-the-art works, some of them are single-criterion models \cite{em:72,em:26} and multi-criteria models \cite{em:4,em:23,em:50, em:2}.
 
We demonstrate the validity of proposed model by comparing results in different encoders with and without HGN. As can be seen from the Table~\ref{t3}, these models with the HGN outperform those baseline models without HGN in terms of F1 value and $R_{oov}$ on almost all datasets, and achieve competitive results on MSR and PKU. We also find the average $R_{oov}$ of RoBERTa-CRF with HGN has been significantly improved comparing with RoBERTa-CRF, nearly 1\% improvement in almost all datasets. These indicate that the HGN can help improve the performance of segmentation and $R_{oov}$.

Among different encoders, the improvement of pre-trained encoders on F1 value is still decent. Compare with using Bi-LSTM as the encoder \cite{em:72,em:4}, the F1 value obtains about 2\% improvement, the $R_{oov}$ gains approximately 10\% when use RoBERTa and BERT as the encoders. The possible reason is that the pre-training process provides some external knowledge.

 Compare to these multi-criteria scenario, although these models trained with more training data, our model achieves better F1 scores and $R_{oov}$. In general, our model effectively alleviates the OOV issue by integrating multi-level linguistic information with HGN.

\begin{table*}
\centering
\caption{ Performance comparison between HGNSeg and previous state-of-the-art models on the test sets of six datasets. And experimental results of HGNSeg on six datasets with different encoders. ``+HGN'' indicates this model uses the heterogeneous graph network. Here, F, $R_{oov}$ represent the F1 value and OOV recall rate respectively. The maximum F1 values are highlighted for each dataset. }            
\begin{tabular}{c|c|c|c|c|c|c|c|c} 
\Xhline{1.2pt}
\hline
\rowcolor{gray!25}
\multicolumn{3}{c|}{\textbf{Models}} & \textbf{AS}                        & \multicolumn{1}{c|}{\textbf{PKU}}               & \multicolumn{1}{c|}{\textbf{MSR}}               & \textbf{CITYU}                      & \multicolumn{1}{c|}{\textbf{CTB}}              & \multicolumn{1}{c}{\textbf{SXU}}                                                       \\ 
\hline
\hline
 \multirow{2}{*}{1} & \multirow{2}{*}{\citet{em:72}} & F   & -          & 95.70                                   & 97.70                                   & -          & 95.95                                  & \multicolumn{1}{c}{-}           \\ 
\cline{3-9}
                &       & $R_{oov}$ & -          & \multicolumn{1}{c|}{-} & \multicolumn{1}{c|}{-} & -          & \multicolumn{1}{c|}{-} & \multicolumn{1}{c}{-}            \\ 
\hline
\hline
\multirow{2}{*}{2} &\multirow{2}{*}{\citet{em:4}}  & F   & \multicolumn{1}{c|}{96.64} & 94.32                                  & 96.04                                  & \multicolumn{1}{c|}{95.55} & 96.18                                  & 96.04                                             \\ 
\cline{3-9}
               &        & $R_{oov}$ & \multicolumn{1}{c|}{72.67} & 72.67                                  & 71.60                                   & \multicolumn{1}{c|}{81.40}  & 82.48                                  & 77.10                                     \\ 
\hline
\hline
\multirow{2}{*}{3}&\multirow{2}{*}{\citet{em:23}} & F  &   -                         & 96.06                                  & 97.25                                  &     -                       & 96.7                                   & 96.47                                   \\ 
\cline{3-9}
           &            & $R_{oov}$ & -          & \multicolumn{1}{c|}{-} & \multicolumn{1}{c|}{-} & -          & \multicolumn{1}{c|}{-} & \multicolumn{1}{c}{-}         \\ 
\hline
\hline
\multirow{2}{*}{4}& \multirow{2}{*}{\citet{em:2}}  & F  & -          & 96.85                                  & 98.29                                  & -          & 97.56                                  & 97.56                                   \\ 
\cline{3-9}
                  &     & $R_{oov}$ & -          & 82.35                                  & 81.75                                  & -          & 88.02                                  & 85.73                                   \\ 
\hline
\hline
\multirow{2}{*}{5}&\multirow{2}{*}{\citet{em:50}} & F   & \multicolumn{1}{c|}{96.44 } &      96.41                          & 98.05                                  & \multicolumn{1}{c|}{96.91} & 96.99                                  & 97.61                                             \\ 
\cline{3-9}
             &          & $R_{oov}$ & \multicolumn{1}{c|}{76.39} & 78.91                                  & 78.92                                  & \multicolumn{1}{c|}{86.91} & 87.00                                     & 85.08                                        \\ 
\hline
\hline
\multirow{2}{*}{6} &\multirow{2}{*}{\citet{em:26}} & F   & \multicolumn{1}{c|}{96.62} & 96.53                                  & 98.40                                   & \multicolumn{1}{c|}{97.93} & 97.25                                  & \multicolumn{1}{c}{-}         \\ 
\cline{3-9}
                    &   &$R_{oov}$ & \multicolumn{1}{c|}{79.64} & 85.36                                 & 84.87                                  & \multicolumn{1}{c|}{90.15} & 88.46                                  & \multicolumn{1}{c}{-}            \\ 
\hline
\hline
\multirow{2}{*}{7} &\multirow{2}{*}{\citet{em:81}} & F   & \multicolumn{1}{c|}{96.86} & \textbf{97.21}                                  & 98.52                                 & \multicolumn{1}{c|}{98.13} & 97.79                                 & \multicolumn{1}{c}{-}         \\ 
\cline{3-9}
                    &   &$R_{oov}$ & \multicolumn{1}{c|}{79.22} & \textbf{90.03}                                  & 86.13                                  & \multicolumn{1}{c|}{91.87} & \textbf{90.15 }                                 & \multicolumn{1}{c}{-}            \\ 
                    
\hline
\hline
\multirow{2}{*}{7} &\multirow{2}{*}{\citet{em:82}} & F   & \multicolumn{1}{c|}{-} & 96.91                              & \textbf{98.69}                                  & \multicolumn{1}{c|}{-} & 97.52                                 & \multicolumn{1}{c}{-}         \\ 
\cline{3-9}
                    &   &$R_{oov}$ & \multicolumn{1}{c|}{-} & -                                 & -                                  & \multicolumn{1}{c|}{-} & -                                 & \multicolumn{1}{c}{-}            \\ 
                    
\hline
\hline
\multirow{2}{*}{8}&\multirow{2}{*}{BERT} &F   & \multicolumn{1}{c|}{96.86} & 96.28                                  &    97.40                                 & \multicolumn{1}{c|}{98.14} & 97.48                                 & \multicolumn{1}{c}{97.32}         \\ 
\cline{3-9}
  & & $R_{oov}$ & \multicolumn{1}{c|}{79.87} & 78.44   &  85.26    & \multicolumn{1}{c|}{91.27} & 89.07                                 & \multicolumn{1}{c}{84.45}          \\ 
\hline
\hline
\multirow{2}{*}{9} &\multirow{2}{*}{BERT+HGN} &F   & \multicolumn{1}{c|}{96.88} &  96.54 &  97.75    & \multicolumn{1}{c|}{98.18} &    97.77& \multicolumn{1}{c}{97.50}         \\ 
\cline{3-9}
  & & $R_{oov}$ & \multicolumn{1}{c|}{80.17} & 80.08     &86.16                               & \multicolumn{1}{c|}{91.47} &  89.67                                & \multicolumn{1}{c}{85.13}          \\ 
\hline
\hline
\multirow{2}{*}{10} & \multirow{2}{*}{RoBERTa} &F   & \multicolumn{1}{c|}{96.73} &  96.68                             &     98.12     &  \multicolumn{1}{c|}{98.29} &  97.54                                & \multicolumn{1}{c}{97.45}        \\ 
\cline{3-9}
  & & $R_{oov}$ & \multicolumn{1}{c|}{78.73} &      79.62                            &87.87        & \multicolumn{1}{c|}{91.47} &   88.61                             & \multicolumn{1}{c}{85.47}         \\ 
\hline
\hline
\multirow{2}{*}{11} &\multirow{2}{*}{RoBERTa+HGN}  & F   &       \textbf{96.88 }                & \multicolumn{1}{c|}{96.81}               & \multicolumn{1}{c|}{98.26}                  &  \textbf{98.36}                          & \multicolumn{1}{c|}{\textbf{97.80}}                  & \multicolumn{1}{c}{\textbf{97.64}}                                                          \\ 
\cline{3-9}
             &          & $R_{oov}$ &         \textbf{80.67 }                  &  \multicolumn{1}{c|}{81.41}                  & \multicolumn{1}{c|}{\textbf{88.39}}                 &   \textbf{91.99}                      & \multicolumn{1}{c|}{89.67}                & \multicolumn{1}{c}{\textbf{86.10}}                   \\

\hline

\Xhline{1.2pt}
\end{tabular}

\label{t3}
\end{table*}

\begin{table*}
\centering
\begin{tabular}{c|c|c|c|c|c|c|c|c} 
\hline
\Xhline{1.2pt}
\multicolumn{1}{c|}{\multirow{2}{*}{\textbf{Models}}} & \multicolumn{2}{c|}{\textbf{DM}}                    & \multicolumn{2}{c|}{\textbf{PT}} 
& \multicolumn{2}{c|}{\textbf{DL}}  
 & \multicolumn{2}{c}{\textbf{ZX}}   \\ 
\cline{2-9}
\multicolumn{1}{c|}{}                        & \multicolumn{1}{c|}{F} & \multicolumn{1}{c|}{$R_{oov}$ } & \multicolumn{1}{c|}{F} & \multicolumn{1}{c|}{$R_{oov}$ } & \multicolumn{1}{c|}{F} & 
\multicolumn{1}{c|}{$R_{oov}$ } & 
\multicolumn{1}{c|}{F} & 
\multicolumn{1}{c}{$R_{oov}$ } \\ 
\hline
\citet{em:77}                                                          & 82.80                  & -                   & 85.00                  & -   & 92.50                 &   -  &   83.90 &  -                                   \\ 
\hline
\citet{em:35}                                                           & 81.20                 & -                      & 85.90                  & -      & 92.00                  & -    & 88.80  &-                                \\ 
\hline
\citet{em:76}                                                             & 82.20                   & -                      & 85.10                  & -   & 93.50                  & -    &  89.60 &  -                                    \\ 
\hline
\citet{em:78}                                                             & 85.00                  & -                      & 89.60                  & -        & \textbf{94.10}                  & -   & \textbf{90.90} & -                                     \\ 
\hline

RoBERTa                  & 89.86                  &   75.91        & 94.06                   & 83.96         & 87.87                 &   62.80  & 86.43  &   76.00                        \\
\hline

Ours                                                            & \textbf{90.31}                    & \textbf{77.40}                     & \textbf{94.38}                   & \textbf{86.92}       & 92.16                  & \textbf{64.50}  & 86.66  &  \textbf{78.71}                             \\
\hline
\Xhline{1.2pt}
\end{tabular}
\caption{The F1 scores and $R_{oov}$ on test data of four cross-domain datasets. The currently maximum values of evaluation are highlighted for each domain dataset.
 }
\label{t4}
\end{table*}

\begin{table*}
\centering
\begin{tabular}{c|c|c|c|c|c|c|c} 
\hline
\rowcolor{gray!25}
\multicolumn{2}{c|}{\textbf{Models}} & \textbf{AS} & \textbf{PKU}   & \textbf{MSR}   & \textbf{CITYU} & \textbf{CTB6} 
& \textbf{SXU}  
\\ 
\hline
\multirow{2}{*}{w/o word/n-gram sub-graph} & F   &  96.76  & 96.64 & 98.09 & 98.08 & \textbf{97.82} & 97.40 
\\ 
\cline{2-8}
                      & $R_{oov}$ &   79.44 & 78.41 & 86.77 & 91.00    & 89.52 & 84.58  
                      \\ 
\hline
\multirow{2}{*}{w/o syntactic sub-graph} & F   &  96.79  & 96.78 &    98.05   &  97.98     &   97.65    &  97.45     
\\ 
\cline{2-8}
                      & $R_{oov}$ &    78.89 & 80.32      &   87.49    &91.07       & 88.8      & 85.15 
                      \\ 
\hline
\multirow{2}{*}{Ours}  & F   &\textbf{96.88}    & \textbf{96.81} & \textbf{98.26} & \textbf{98.36} & 97.80  & \textbf{97.64} 
\\ 
\cline{2-8}
                      & $R_{oov}$ & \textbf{80.67}   & \textbf{81.41}  & \textbf{88.39} & \textbf{91.99} & \textbf{89.67}  & \textbf{86.10}  
                      \\
\hline

\end{tabular}

\caption{Ablation experiments, ``w/o word/n-gram sub-graph'' means without the character-word/n-gram sub-graph, ``w/o Syntactic sub-graph'' means without the dependency syntactic sub-graph. }
\label{t5}
\end{table*}

\subsection{Cross-Domain Performance}

We also compare our model with previous methods for cross-domain CWS, the main results are listed in Table~\ref{t4}. We use words from PKU training set and n-grams from each cross-domain corpus to build the character-word/n-gram sub-graph. According to the results in Table~\ref{t4}, although we didn't use external domain dictionary, our model outperforms four baselines for DM, PT. But our model performed unsatisfactorily on two fantasy novel datasets, the one possible reason is these two datasets have different segmentation criterion with PKU, another reason is there is a big gap between novel text and news text. Specifically, compare to four previous works, the F1 values of DM, PT obtain a 5\% average improvement. We also can see that our model with HGN boost the $R_{oov}$ on four datasets significantly compare with only using RoBERTa, the $R_{oov}$ on DM, PT, DL, ZX improve 1.49\%, 2.96\%, 1.7\%, 2.71\% respectively. 

\begin{figure}
\centering
\includegraphics*[width=0.5\textwidth]{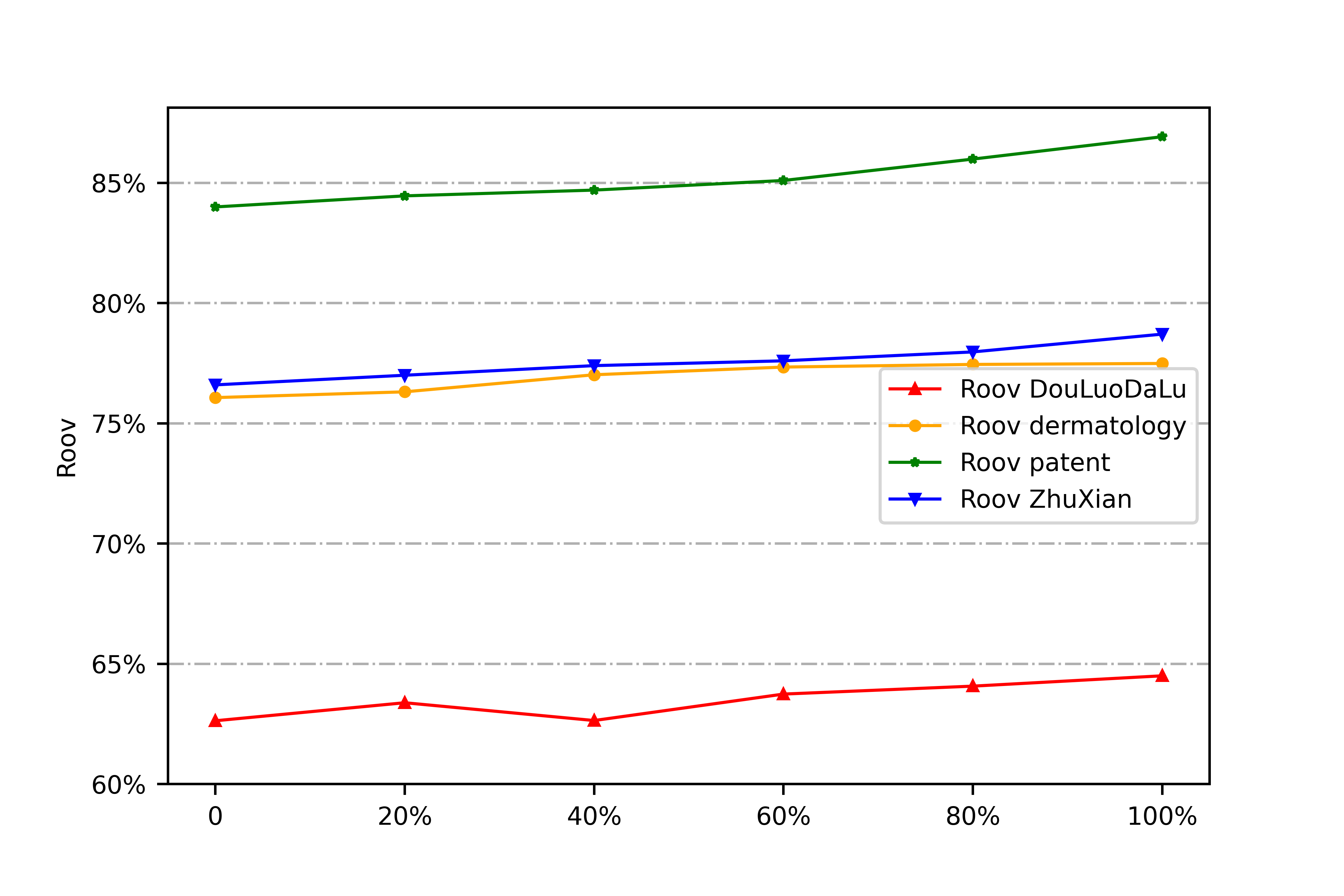}
\caption{$R_{oov}$ with different sizes n-gram vocabulary for four cross-domain datasets. We randomly select different proportions of n-grams from the each target corpora to generate new vocabularies}
\label{f4}
\end{figure}

As we know, cross-domain CWS is exposed to a major challenge, OOV problem. Segmenters built on the newswire domain have very limited information to segment domain-specific words. Since we also investigate the influence of the domain-specific n-grams vocabulary size on $R_{oov}$. We randomly selected 20\%, 40\%, 60\% and 80\% of the n-grams from each target corpus to construct new n-grams vocabulary with different sizes. Figure~\ref{f4} shows the $R_{oov}$ of our model with these vocabularies. From Figure~\ref{f4}, we can see the performance of our model improves gradually with the vocabulary size increases. Therefore, we could infer that our model will get better results if we can obtain a vocabulary containing more domain-specific n-grams. The $R_{oov}$ on PT gets the biggest improvement. By analyzing some cases in PT test set, we found that our model can recognize some domain-specific words, such as ``蛋白激酶(Protein kinase)'', ``聚丙烯蜡(Polypropylene wax)'', ``白芥子(Mustard Seeds)'', ``酶制剂(Enzyme)'' et al.. These indicate that domain-specific n-grams can help improve the $R_{oov}$ effectively.

\subsection{Ablation Study}
In this section, we discuss the effectiveness of each component in the heterogeneous graph neural network. The experimental results show in Table~\ref{t5}.

The first ablation study is to evaluate the effect of the character-word/n-gram sub-graph. The comparison between the first and the third line in Table~\ref{t5} indicates that this type of sub-graph boost the $R_{oov}$ on six datasets, all datasets obtain about 1\% improvement, especially PKU and MSR are more sensitive to this type of sub-graph, obtain 3\% and 1.62\% improvement respectively.

The second ablation study is to verify the effectiveness of the dependency syntactic sub-graph. In this experiment, we test the performance of HGNSeg by removing the dependency syntactic sub-graph from the heterogeneous graph. As we can see the second line in Table~\ref{t5}, although this type of sub-graph also boosts the overall F1 score and $R_{oov}$, it seems that majority datasets are more sensitive to the character-word/n-gram sub-graph.




\subsection{Effect of Different Lexicons}
We also analyze the validity of different lexicons in the character-word/n-gram sub-graph, i.e., with and without lexicon from training set. In the character-word/n-gram sub-graph, there are two types of lexicon, one is based on training set, another one is based on high frequency n-grams from each corpus. The experimental results with different lexicons show in Table~\ref{t6}. We can see that their effects vary in different datasets. In terms of average performance, the lexicon from training set and n-grams vocabulary both are important and boost the performance considerably. 
\begin{table*}[ht]
\centering
\begin{tabular}{c|c|c|c|c|c|c|c} 
\hline
\Xhline{1.2pt}
\rowcolor{gray!25}
\multicolumn{2}{c|}{\textbf{Models}}      & \textbf{AS}    & \textbf{PKU}                                  & \textbf{MSR}                                  & \textbf{CITYU} & \textbf{CTB6}  & \textbf{SXU}  
\\ 
\hline
\multirow{2}{*}{w/o $\mathcal{N}$} & F   &  96.84     & 96.68                                & 98.22                                & 98.27 & 97.65 & 97.36
\\ 
\cline{2-8}
                            & $R_{oov}$ &  78.80     & 79.59                                & 87.92                                & 90.72 & 89.00    & 85.02 
                            \\ 
\hline
\multirow{2}{*}{w/o $\mathcal{D}$} & F   &   \textbf{96.89 }   & 95.03& 96.65&98.14       &   97.67    &      97.21
\\ 
\cline{2-8}
                            & $R_{oov}$ &   80.28    & 79.26 & 86.59  &     90.40  &    89.00   &  85.54    
                            \\ 
\hline
\multirow{2}{*}{Ours}       & F   & 96.88 & \textbf{96.81}                                & \textbf{98.26}                                & \textbf{98.36} & \textbf{97.80} & \textbf{97.64} 
\\ 
\cline{2-8}
                            & $R_{oov}$ & \textbf{80.67} & \textbf{81.41}                                & \textbf{88.39}                                & \textbf{91.99} & \textbf{89.67} & \textbf{86.10} 
                            \\
\hline
\Xhline{1.2pt}
\end{tabular}
\caption{Comparison of different types of lexicon in the character-word/n-gram sub-graph. ``w/o $\mathcal{N}$ '' means without the n-grams vocabulary, ``w/o $\mathcal{D}$'' means without the lexicon from training set. Last line represents using $\mathcal{N}$ and $\mathcal{D}$ together.}
\label{t6}
\end{table*}

\subsection{Effect of Different Syntactic Parsing Toolkits}

\begin{figure}
\centering
\includegraphics*[width=0.5\textwidth]{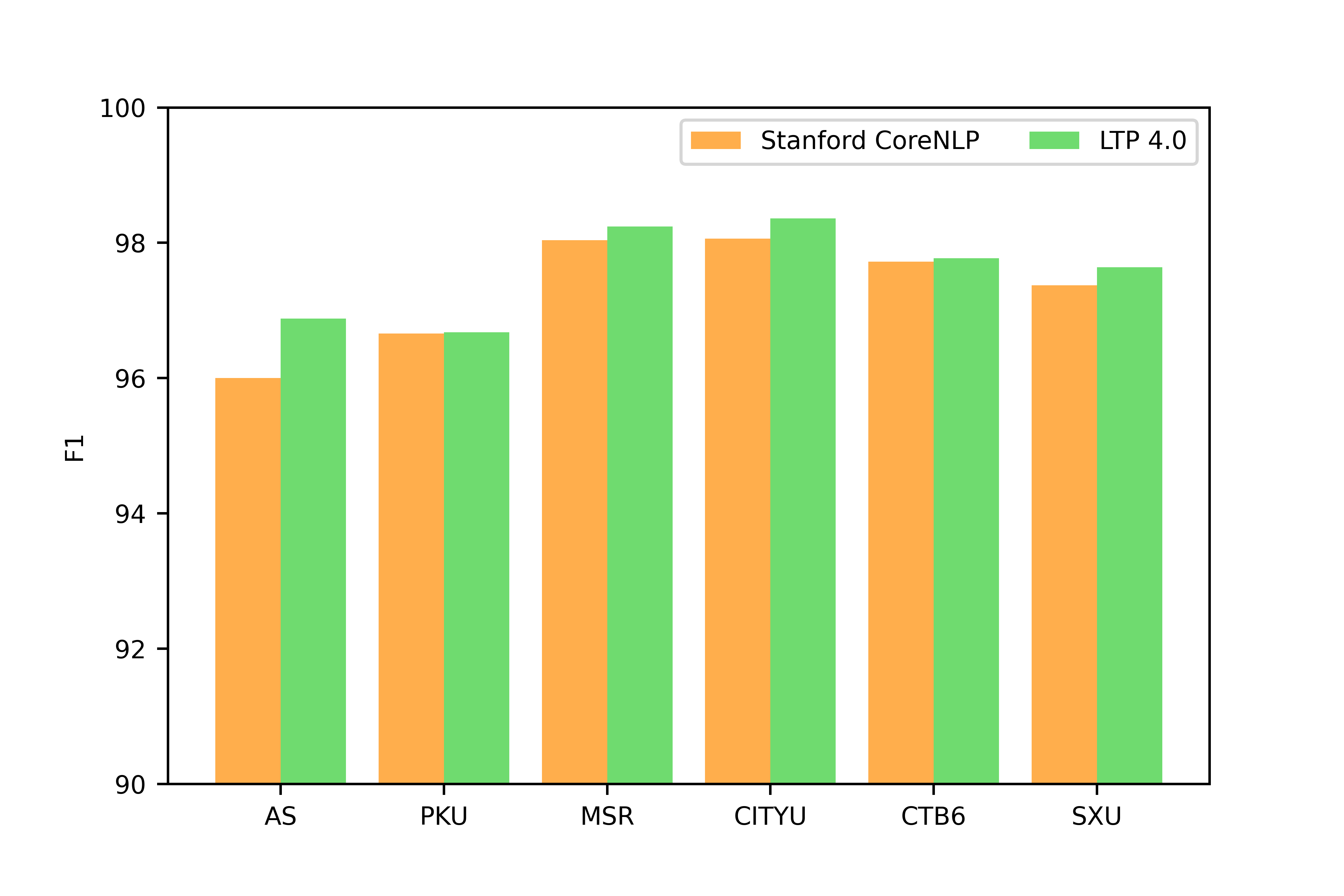}
\caption{The F1 values of HGNSeg using two different dependency syntactic parsing toolkits, namely Stanford CoreNLP Toolkit and LTP 4.0, on six datasets.}
\label{f3}
\end{figure}

Through the previous analysis, our model can incorporate syntactic information effectively. However, existing syntactic parsing toolkits are not perfect, there are always some parsing mistakes, particularly when use on the long sentence. Since we compare the effects of two dependency syntactic parsing toolkits. We draw the histograms of the F1 scores obtain from HGNSeg with different parsing methods on six datasets (yellow bars for Stanford CoreNLP Toolkit, green bars for LTP 4.0) in Figure~\ref{f3}.

According to the figure, we can see LTP4.0 is more suitable for this task, and the performance of CWS on six datasets is better than Stanford CoreNLP. Although Stanford CoreNLP toolkit provides rich dependency syntactic labeling information, we only use the head and dependent information. Therefore, the dependency syntactic parsing of LTP 4.0 is sufficient to meet the needs of the CWS task in our framework. In the future, we will consider incorporating more dependency labeling information from Stanford CoreNLP to improve CWS.

\section{Conclusion}

In this work, we proposed a novel framework, HGNSeg, for CWS using multi-level linguistic information. We use the heterogeneous graph neural network to encode the structure information of words, n-grams, and dependency syntax. To the best of our knowledge, this is the first work using the heterogeneous graph neural network for neural models in CWS. Experimental results on six benchmark datasets illustrate the effectiveness of our model. Further experiments and analysis also demonstrate the robustness of HGNSeg in the cross-domain scenario. 
\bibliography{anthology,custom}
\bibliographystyle{acl_natbib}

\appendix


\end{CJK*}
\end{document}